\documentclass[
twocolumn, 
hf, 
]{ceurart}

\usepackage{listings}
\usepackage{float}
\usepackage{marginnote}
\usepackage{footnote}
\usepackage{multirow}

\usepackage{amsmath}
\usepackage{mathtools}
\usepackage{graphicx}
\usepackage{subcaption}

\begin{document}

\conference{\href{https://hexed-workscopyhop.github.io}{HEXED'25: 2nd Human-Centric eXplainable AI in Education Workshop}, 20 July, 2025, Palermo, Italy}

\title{RegKT: Interpretable and Robust Deep Knowledge Tracing With IRT-Regularizer}

\author[1]{Samuel Girard}[%
orcid=0009-0009-8205-7484,
email=samuel.girard@inria.fr]
\cormark[1] 
\address[1]{Inria-Saclay}

\author[2]{Juan D. Pinto}[%
    orcid=0000-0002-2972-485X,
    email=jdpinto2@illinois.edu,
    url=https://jdpinto.com,]
\address[2]{University of Illinois Urbana-Champaign, Urbana, IL, USA}

\author[1]{Jill-Jênn Vie}[%
orcid=0000-0002-9304-2220,
email=jill-jenn.vie@inria.fr,
url=https://jill-jenn.net,
]

\author[3]{Amel Bouzeghoub}[%
orcid=0000-0003-4890-9005,
email=amel.bouzeghoub@telecom-sudparis.eu,
url=https://amel.wp.imtbs-tsp.eu,
]
\address[3]{Telecom-Sud Paris}

\cortext[1]{Corresponding author.}

\begin{abstract}
As deep learning models continue to advance, knowledge tracing models have achieved higher accuracy. However, these gains come at the cost of reduced interpretability, which is crucial for practitioners in educational settings to adopt new methodologies. Additionally, deep learning models are prone to overfitting, particularly when dealing with the small datasets that are common in educational applications. In this paper, we propose a novel regularization technique designed to enhance the robustness of deep-learning-based knowledge tracing models, while simultaneously improving their interpretability. Our method addresses both the interpretability and overfitting challenges, making it more feasible for real-world educational applications.
\end{abstract}


\maketitle

\section{Introduction}

Deep learning (DL) has significantly advanced the field of knowledge tracing, enabling more accurate predictions of student performance over time. Knowledge tracing models have evolved from early methods, such as Bayesian knowledge tracing (BKT), to more sophisticated approaches like deep knowledge tracing (DKT), which leverage the power of recurrent neural networks. These advancements have led to substantial improvements in predictive accuracy, making knowledge tracing models indispensable tools for adaptive learning platforms and personalized education.

Yet, despite these improvements, DL-based knowledge tracing models suffer from a critical limitation: they often lack interpretability \cite{dingWhyDeepKnowledge2019, dingInterpretabilityDeepLearning2021, huangxkt}. That is, while they can accurately predict students' success on the next problem, they lack the interpretable parameters of BKT or item response theory (IRT). In educational settings, where teachers and instructional designers rely on clear, understandable results, the opaque nature of deep models limits their adoption \cite{lu2020interpretabledeeplearningmodels, pintoDeepLearningEducational2024}. This also makes such models unusable for tasks such as open learner modeling \cite{kayEnhancingLearningOpen2022}. Additionally, DL models are susceptible to overfitting, especially in contexts where educational datasets are small or sparse \cite{Gervet_Koedinger_Schneider_Mitchell_2020}.

To address these issues, we introduce a novel hybrid model, RegKT, which integrates some of the interpretability of IRT with the temporal modeling capabilities of DKT \cite{sun2024, zhang2024questioncentricmultiexpertscontrastivelearning}. By incorporating a regularization term based on IRT into the DKT framework, we can control the trade-off between accuracy and interpretability using a hyper-parameter \( \epsilon \). This hybrid approach allows our model to achieve high accuracy without sacrificing explainability, making it a better fit for real-world educational applications.

The remainder of this paper is structured as follows: we first review related work on interpretability in DL-based knowledge tracing models and regularization techniques for educational applications. We then describe our methods and present the results of our approach. We also touch on the implications that our work has on interpretability. Finally, we discuss future research directions.

\section{Related Work}

The task of estimating students' latent knowledge states is one that has been studied at depth using a variety of approaches. In this section we review some past approaches that are relevant to our integrated RegKT model.

\subsection{Item Response Theory}

IRT \cite{raschProbabilisticModelsIntelligence1993} is one approach based on logistic regression and used primarily in the field of assessment and psychometrics. IRT models the relationship between a student's latent ability and their performance on test items, making it highly interpretable and less prone to overfitting compared to more complex models.

The basic IRT model, known as the one-parameter logistic model or Rasch model, is given by:

\[
P(X_{ij} = 1 | \theta_i, b_j) = \frac{1}{1 + \exp{-(\theta_i - b_j)}}
\]

Where \( P(X_{ij} = 1) \) represents the probability that student \( i \) correctly answers item \( j \), \( \theta_i \) denotes the latent ability of student \( i \), and \( b_j \) is the difficulty parameter of item \( j \).

IRT produces simple, interpretable curves that illustrate the probability of a correct response based on student ability. This framework's simplicity helps prevent overfitting, allowing it to generalize well, even with smaller datasets. However, its limitation is that it assumes a student's ability is static, whereas knowledge evolves over time.



\subsection{Deep Knowledge Tracing}

Knowledge tracing, as another family of methods, emphasizes knowledge growth over time—something IRT does not account for. The most well-known models for knowledge tracing include BKT \cite{corbettKnowledgeTracingModeling1995}, a hidden Markov model approach, and DKT \cite{piech2015deepknowledgetracing}, built on recurrent neural networks and their long short-term memory (LSTM) variants.

DKT and other DL-based knowledge tracing algorithms have become popular in the literature due to their improvements in accuracy and their lack of reliance on predefined assumptions about knowledge structure.

In DKT, for each student interaction, the model updates its hidden state \( h_t \), which encodes the student's evolving knowledge. The model's forward pass can be described as:

\begin{align}
    h_t &= \tanh(W_h^{hx} x_t + W_h^{hh} h_{t-1} + b_h) \\
    y_t &= \sigma(W_y^h h_t + b_y)
\end{align}



DKT's flexibility comes from its ability to learn complex patterns directly from student interaction data. However, this flexibility comes at the cost of interpretability due to the interplay between flexibility, model complexity, and interpretability. Furthermore, DKT models are prone to overfitting, particularly when training on small datasets.

\subsection{Integrating Approaches}

To our knowledge, \cite{yeungDeepIRTMakeDeep2019} is the only prior work that attempts to integrate the advantages of IRT and DKT into a single model. Rather than the original DKT, they rely on the dynamic key-value memory network (DKVMN) variant \cite{zhangDynamicKeyvalueMemory2017}, which creates a memory matrix that maps student knowledge states to implicitly derived knowledge components (though the accuracy of this implicit mapping and the interpretability of such an approach has been previously questioned---see \cite{dingInterpretabilityDeepLearning2021}). They use this matrix's vector representations to infer student ability and difficulty level, which they then plug into the IRT model as $\theta$ and $b$ to predict student success on the next item. While their model obtains promising results, their reliance on DKVMN increases the data requirements of such an approach.

Our approach differs from this in that we aim to constrain DKT via an added penalty term to its loss function. A similar constraints-based approach was followed by \cite{pintoInterpretableNeuralNetworks2023}, in which a convolutional neural network trained to predict gaming-the-system behavior was constrained in order for the network to learn binary rather than continuous weights in its convolutional layer. As with our approach, the researchers achieved this using an added regularization term to their model's loss function. Their rationale for binarizing weights was that it would allow the model to better fit the nature of the features with the goal of improving interpretability.

Similarly, \cite{shiInterpretableCodeinformedLearning2023} proposed introducing a penalty term to the loss function of a DKT model to ensure that the knowledge components learned would follow sensible learning curves. However, this was a theoretical proposal, so the details and results of the implementation are not available.

While there are similarities between our approach and these prior works, our method is unique in that it seeks to integrate specific interpretable parameters from a different model with the temporal modeling capabilities of DKT. This allows us to maintain some of the interpretability of IRT while benefiting from the flexibility of DKT.

\section{Methodology}

\begin{figure*}[h]
    \centering
    \includegraphics[width=0.5\textwidth]{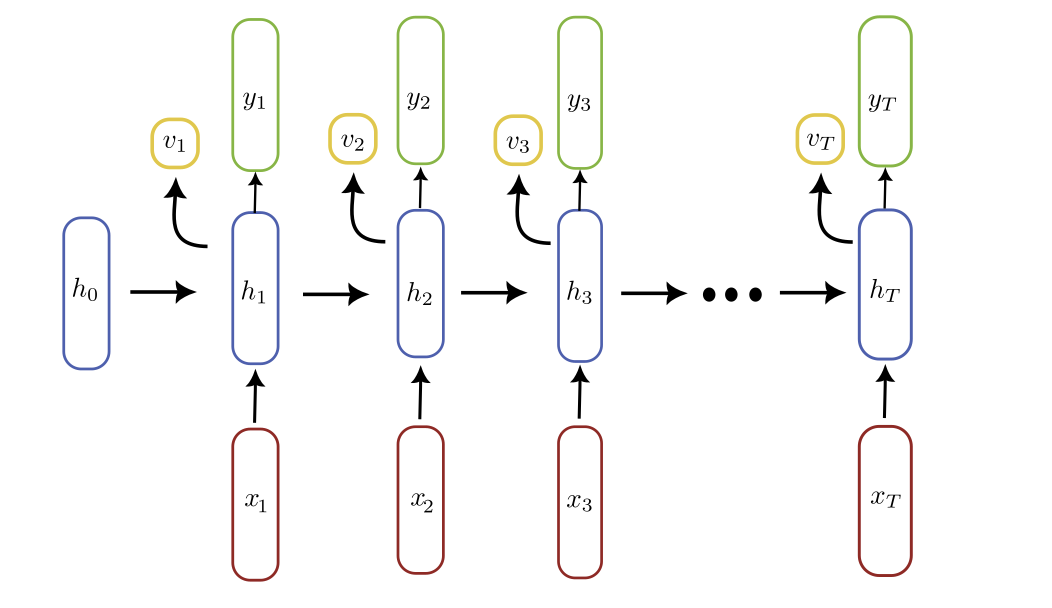}  
    \caption{The connection between variables in RegKT. The input ($x_t$) is either a one-hot encoding or compressed representation of student action, and the prediction ($y_t$) is a probability vector for answering each problem correctly, with $(v_t)$ as the student's proficiency parameter estimate.}
    \label{fig:wide-figure}
\end{figure*}

To address the limitations of both IRT and DKT, we propose a hybrid model, RegKT, which integrates the key properties of both approaches. RegKT uses DKT's temporal modeling capabilities while incorporating an IRT-based regularization term to enhance interpretability.

\subsection{Model}

Traditional recurrent neural networks are designed to map a sequence of input vectors \( x_1, \ldots, x_T \) to a corresponding sequence of output vectors \( y_1, \ldots, y_T \). This is accomplished by maintaining a sequence of hidden states \( h_1, \ldots, h_T \), where each hidden state \( h_t \) encodes relevant information from previous time steps t. Essentially, the hidden state \( h_t \) represents the student’s evolving knowledge over time as they interact with items. In RegKT, we do not only have one sequence of output vectors but also a sequence of scalar outputs \( v_1, \ldots, v_T \) that correspond to the prediction of the latent ability parameter of the given student.
The variables are related using a simple network defined by the equations:

\begin{align}
    h_t &= \tanh(W_h^{hx} x_t + W_h^{hh} h_{t-1} + b_h) \\
    y_t &= \sigma(W_y^h h_t + b_y)\\
    v_t &= W_v^h h_t + b_v
\end{align}

Equations (3) and (4) are exactly the same as in DKT and equation (5) is a simple perceptron linking the hidden state to the  latent ability estimation.
The architecture of the model can be visualized in figure \ref{fig:wide-figure}.

As with DKT, the input  $x_t$ is the  one-hot encoding of the student interaction tuple ${q_t,a_t}$ that represents the combination of which exercise was answered and if the exercise was answered correctly, so $x_t \in \{0, 1\}^{2M}$ with M the number of exercise in the corpus.

\subsection{Training Objective}

The training is the most important part of our model as it is here that it differs from DKT. We first start by splitting the dataset into a train set $\mathcal{D}_{train}$ and test set $\mathcal{D}_{test}$ by students i.e no student can appear in both the train set and test set.

From this we fit a simple IRT-1PL model with only $\mathcal{D}_{train}$ to estimate the ability parameter $\theta_i$ of every student $i$ in the train set. The values $\theta$ will serve as a target for the predictions \( v_1, \ldots, v_T \) during the training procedure.

The training objective is the negative log likelihood of the observed sequence of student responses under the model. Let \( \delta(q_{t+1}) \) be the one-hot encoding of which exercise is answered at time \( t + 1 \), and let \( l \) be binary cross entropy. The loss for a given prediction is \( l(y^T \delta(q_{t+1}), a_{t+1}) \), and the loss for a single student $i$ is:
\[
L_i = \underbrace{\sum_t^{T_i} l(y^T \delta(q_{t+1}), a_{t+1})}_{\mathclap{\text{DKT loss objective}}}
+ \underbrace{\epsilon \cdot L_{\text{irt}}\left(f\left((v_j)_{j \leq T_i}\right), \theta_i\right)}_{\mathclap{\text{IRT-based regularization term}}}
\]
Here $f(.)$ is a simple linear form that maps the vector $(v_j)_{j \leq T_i}$ into a scalar (we chose to use $f((v_j)_{j \leq T_i})=v_{T_i}$ as our last prediction should have converged better to fixed parameter $\theta_i$) and $L_{\text{irt}}$ is  a classical error signal such as the MSE or the L1 distance.

The key innovation in RegKT is the introduction of a regularization term based on IRT. This term ensures that the model's predictions align with the IRT-based estimates of student ability \( \theta_i \).

We hypothesize that this regulator allows for more interpretable weights during gradient descent:
the gradient loss from $L_{\text{irt}}$ is backpropagated into the weights of the RNN that are used in the calculation of the hidden states \( h_1, \ldots, h_T \) and the prediction sequences  \( y_1, \ldots, y_T \).

This allows the hidden states to be more explainable and to carry less noise in the data---since they are inputs of a simple 1-layer perceptron they cannot vary too much as $v_{T_i}$ should converge toward $\theta_i$.

Additionally, the scaling weight \( \epsilon \) allows for interpolation between weights in the RNN that are optimized to reduce $L_{\text{irt}}$ and weights that are optimized to capture more complex patterns in the data.
This is useful in a setting where the data is scarce, as it implies that we should sometime put more trust in the loss signal of an underfitted model rather than a complex one that could capture unwanted noise.

\section{Results}

In this section, we present two sets of results that demonstrate the effectiveness of our proposed method.
The first set of results centers on the model's explainability. We provide visual interpretations of the learned weights. These visualizations can be valuable for practitioners, as they make the model's decision-making process more transparent and accessible, potentially guiding instructional decisions. The second set focuses on the model's ability to mitigate overfitting, particularly in small dataset scenarios. By comparing performance across different dataset sizes, we show that our approach can achieve better or similar accuracy than existing methods when data is limited, highlighting its robustness.


To provide more technical details, the results in the table \ref{tab:vertical_table} are from a grid search to find the best hyperparameters.

\begin{table}[h]
    \centering
    \begin{tabular}{|c|c|c|c|}
        \hline
        \textbf{Dataset} & \textbf{Model} & \textbf{AUC} & \textbf{Accuracy} \\
        \hline
        \multirow{3}{*}{\textbf{Fraction}} & DKT & 0.879 & 0.805 \\
        & RegKT & 0.889 & 0.812 \\
        & IRT & 0.656 & 0.615 \\
        \hline
        \multirow{3}{*}{\textbf{RoboMission}} & DKT & 0.835 & 0.898 \\
        & RegKT & 0.838 & 0.9015 \\
        & IRT & 0.786 & 0.890 \\
        \hline
        \multirow{3}{*}{\textbf{ASSISTments2009}} & DKT & 0.741 & 0.696 \\
        & RegKT & 0.772 & 0.723 \\
        & IRT & 0.666 & 0.634 \\
        \hline
        \multirow{3}{*}{\textbf{Synthetic BKT}} & DKT &0.708 &  0.731 \\
        & RegKT & 0.727 & 0.743 \\
        & IRT & 0.681 &  0.719 \\
        \hline
        \multirow{3}{*}{\textbf{Synthetic M-IRT}} & DKT &0.712 &   0.664 \\
        & RegKT & 0.732 & 0.677 \\
        & IRT & 0.612 &  0.584 \\
        \hline
    \end{tabular}
    \caption{Results in AUC and accuracy (Acc.) for all datasets.}
    \label{tab:vertical_table}
\end{table}

\subsection{Fractions}

Fractions is a very simple dataset consisting of 535 users answering 20 exercises on fractions. In our setup, we want a model with low complexity to allow visualization hence our choice to have,for both classical DKT and RegKT, the hidden state of dimension 2. We also chose the number of layers in our LSTM to be 5 (the number of LSTM units stacked together) as it gave better accuracy.

\begin{figure}
    \centering
    \begin{minipage}{0.38\textwidth}
        \centering
        \includegraphics[width=0.8\linewidth]{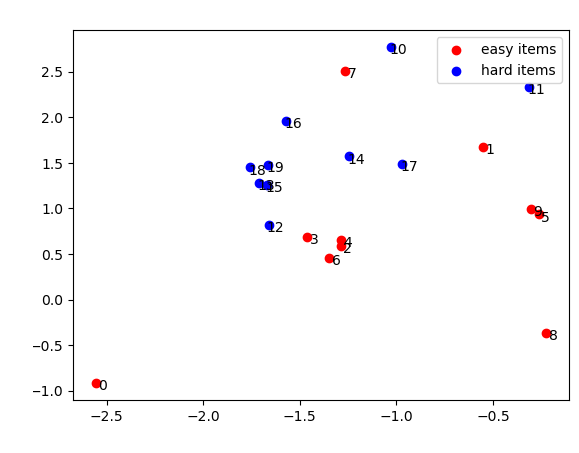}
        \vspace{2pt}
        \parbox{0.8\linewidth}{\centering\small In classical DKT}
    \end{minipage}
    \hfill
    \begin{minipage}{0.38\textwidth}
        \centering
        \includegraphics[width=0.8\linewidth]{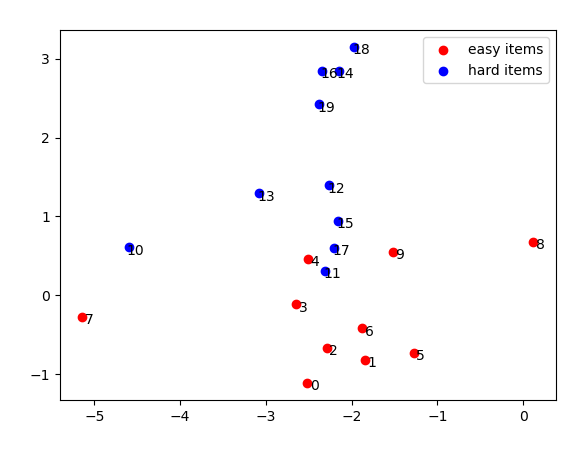}
        \vspace{2pt}
        \parbox{0.8\linewidth}{\centering\small In our model (RegKT)}
    \end{minipage}

    \caption{Visualization of the weights linking the hidden state to item correctness, with each label associated with the ordering of difficulties.}
    \label{fig:two_figures_same_column}
\end{figure}

In figure \ref{fig:two_figures_same_column} we observe that our model allows for better clusters of items.

\begin{figure}
    \centering
    \begin{minipage}{0.38\textwidth}
        \centering
        \includegraphics[width=0.8\linewidth]{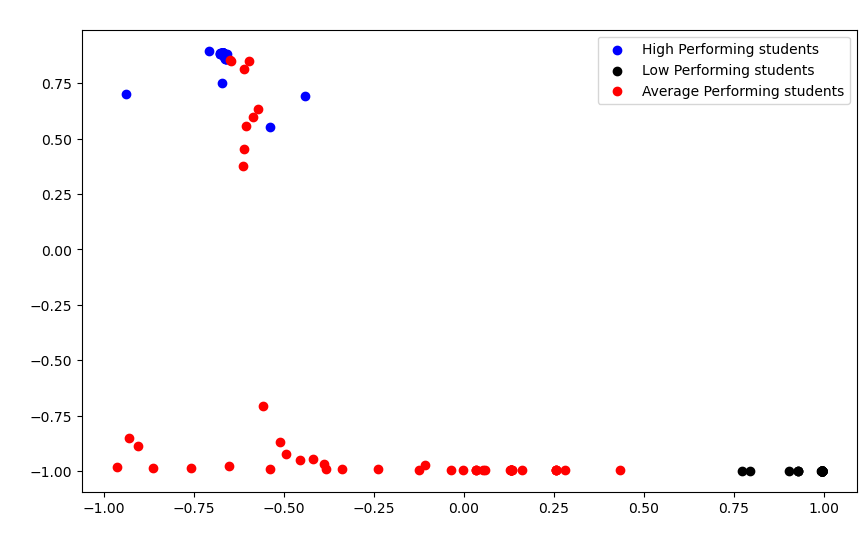}
        \vspace{2pt}
        \parbox{0.8\linewidth}{\centering\small In classical DKT}
    \end{minipage}
    \hfill
    \begin{minipage}{0.38\textwidth}
        \centering
        \includegraphics[width=0.8\linewidth]{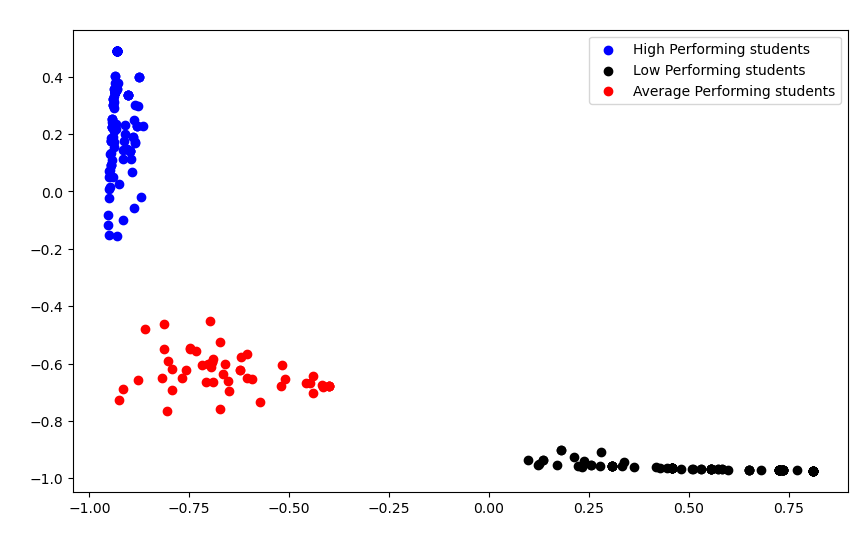}
        \vspace{2pt}
        \parbox{0.8\linewidth}{\centering\small In our model (RegKT)}
    \end{minipage}

    \caption{Visualization of the hidden state clusters in classical DKT (top) vs. RegKT (bottom) in fraction.
    We select subsets of top, middle, and low-performing students (based on their IRT ability) }
    \label{fig:hidden_state_frac}
\end{figure}

We have also looked at the last hidden state $h_{T}\in \mathbb{R}^2$ for every student, which is essential for practitioners to interpret, as it encodes the student's knowledge derived from the entirety of their learning trace. Before running the experiment, we calculated the student's proficiency parameter using a simple 1-PL IRT model with the full dataset. This allowed us to better understand the visualizations, as we noticed that the clusters of the hidden states aligns with the results from IRT (see figure \ref{fig:hidden_state_frac}).

We notice that our method allows for much better  clusters that makes more use of the 2-D plane (Figure \ref{fig:hidden_state_frac}), hence giving more room for interpretation between different types of knowledge states.



\subsection{RoboMission}

The RoboMission dataset consists of over 20,000 students, each answering questions in a corpus of 85 exercises. In this larger-scale setting, we maintained model interpretability by using classical DKT and RegKT models with a hidden state dimension of 2 and 5  layers.

Similarly, we analyzed the final hidden state $h_{T_i} \in \mathbb{R}^2$ for student $i$.


\begin{figure}
    \centering
    \begin{minipage}{0.38\textwidth}
        \centering
        \includegraphics[width=0.8\linewidth]{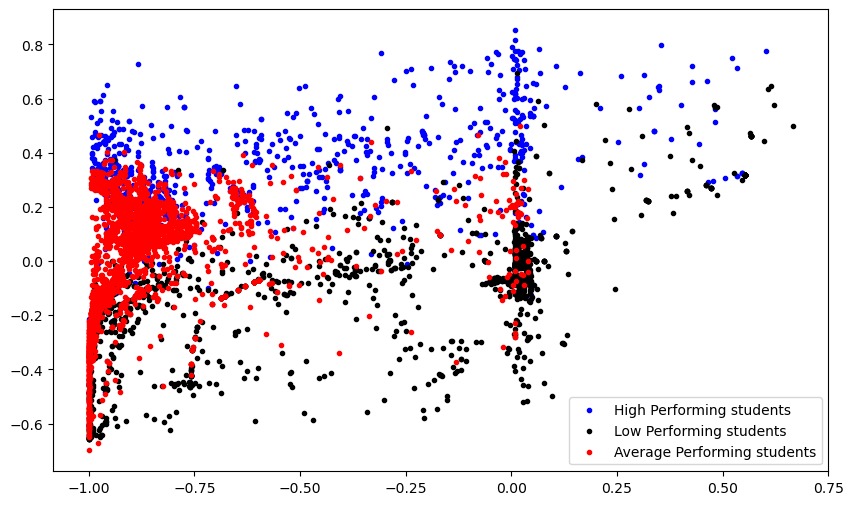}
        \vspace{2pt}
        \parbox{0.8\linewidth}{\centering\small In classical DKT}
    \end{minipage}
    \hfill
    \begin{minipage}{0.38\textwidth}
        \centering
        \includegraphics[width=0.8\linewidth]{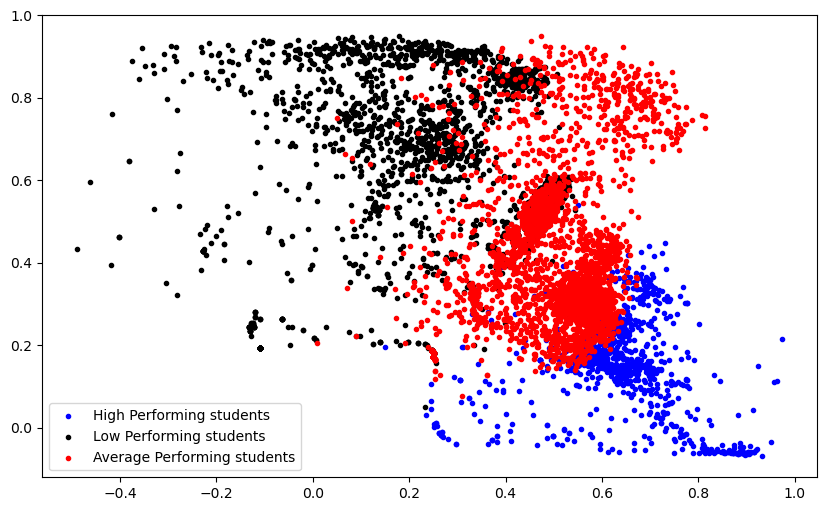}
        \vspace{2pt}
        \parbox{0.8\linewidth}{\centering\small In our model (RegKT)}
        \label{rb2A}
    \end{minipage}

    \caption{Visualization of the hidden-state clusters in classical DKT (top) vs. RegKT (bottom) in RoboMission.}
    \label{fig:hidden_state_eob}
\end{figure}

Figure \ref{fig:hidden_state_eob} shows that our method leads to more distinct clusters in the 2-D projection of the hidden states, but there are no better separations where it comes to the visualizations of the weights linking to item correctness $W^h_y$. Concurrently we do not observe considerable improvement in accuracy.

\subsection{ASSISTments}

We used the ASSISTments 2009-2010 dataset, where we have only looked at the 500 most played items with about 2675 distinct users. In this data set, we have had better accuracy with $h_t\in \mathbb{R}^5$ as it allows us to capture more complex patterns in the data. In order to allow visualizations, we decide to run a PCA with two components on the hidden states.

\begin{figure}[h]
    \centering
    \includegraphics[width=0.38\textwidth]{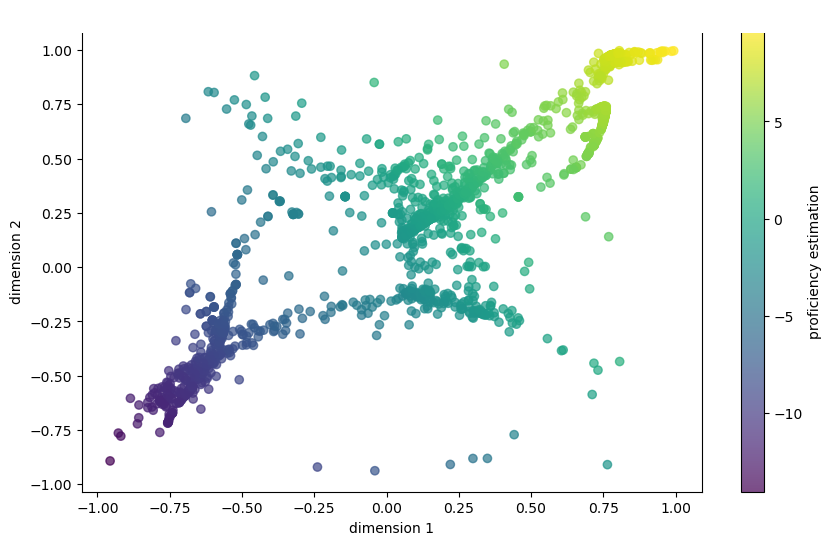}

    \caption{Alignment of the hidden state with its proficiency parameter in ASSISTments: $h_{T_i}$ associated to higher proficiency parameter $v_{T_i}$ are on the top right of the plot}
    \label{fig:hidden_state_alignment}
\end{figure}





\subsection{Synthetic Datasets}

To better understand the benefits of RegKT, we tested our model on synthetic data where we could control the underlying dimensionality. In our synthetic M-IRT experiments, we generated learning traces for 200 students using a corpus containing a variable number \(K\) of exercises and a model with dimensionality \(D\). In this setting, each student is characterized by a \(D\)-dimensional ability vector, while each item is defined by a \(D\)-dimensional discrimination vector and an associated difficulty parameter. Each student responds to a random subset of available items.

\begin{figure}[h]  
    \centering
    \includegraphics[width=1\columnwidth]{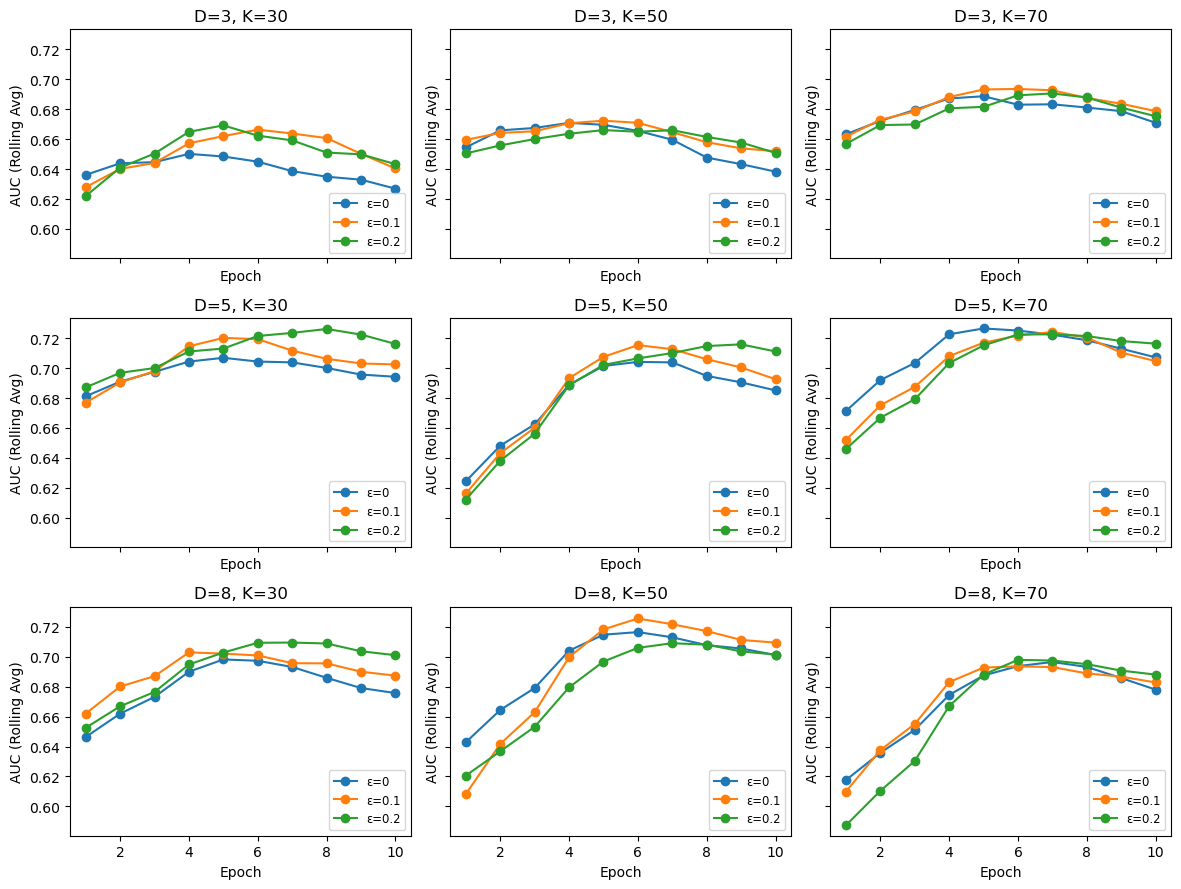}
    \caption{Illustration of the AUC curve, where students interact with exercises generated under the M-IRT model. Here $\epsilon=0$ implies the DKT model}
    \label{fig:synthetic_irt}
\end{figure}

RegKT outperforms DKT when the data is generated from a simpler underlying model, characterized by lower values of $K$ and $D$. This suggests that high-capacity deep learning models, like DKT, tend to capture not only meaningful patterns but also more noise when no regularization is applied. In contrast, RegKT appears to mitigate this issue, leading to better generalization in lower-complexity settings, as illustrated in figure \ref{fig:synthetic_irt}.

Experiments have also been made with a synthetic dataset with learning traces generated from the BKT-student model.

\section{Enhancing interpretability}

The diagrams above demonstrate how our approach improves the clustering of hidden states, making it easier for practitioners to understand how the model makes decisions. Clearer clusters help identify states that correspond to similar knowledge levels or learning patterns, allowing for a more interpretable analysis of a student's evolving ``knowledge state'' over time. This provides insights into how the model perceives mastery progression after each question.

However, our approach's most important contribution to enhancing interpretability lies in its association of each hidden state $h_{t_i}$ with a proficiency scalar parameter $v_{t_i}$. This makes it possible to use students' proficiency estimates in ways similar to more traditional KT algorithms. For example, we can calculate learning curves \cite{vandesandeLearningCurvesProblems2016}, which allow us to determine whether the model perceives the student as improving. Such curves can be used to test the whether the model's understanding of a student is sound or to identify potential issues in the model.

\begin{figure}[h]  
    \centering
    \includegraphics[width=0.8\columnwidth]{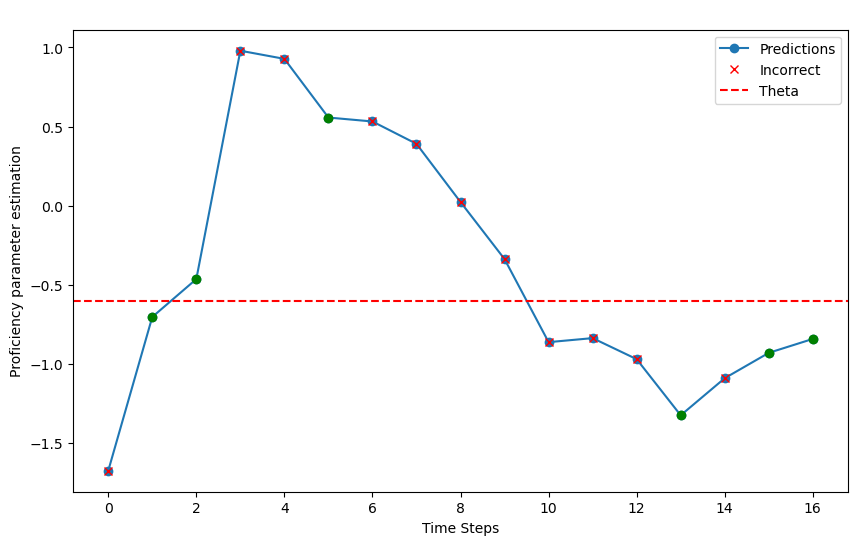}
    \caption{Evolution of the proficiency parameter for a student in the test set. The red dotted line represents a ground-truth proficiency parameter for the student}
    \label{fig:evol_proficiency_test}
\end{figure}

We can observe in figure \ref{fig:evol_proficiency_test} that both outputs of our model align properly. Notably, the IRT estimation in RegKT converges well toward the ground truth\footnote{There is no actual parameter theta. This is simply an estimation calculated on the full dataset.}
proficiency parameter (red dotted line), even for students in the test set, reinforcing the model's reliability in assessing student learning.

Students' proficiency parameter can also be used to create dashboard visualizations for teachers and students, providing an explanation for their learning trajectory \cite{hooshyarOpenLearnerModels2020}. Deriving such explanations directly from RegKT's parameters provides certain advantages over post-hoc explainability approaches that make indirect predictions of the model's internal workings \cite{liuApplicationsExplainableAI2024a}.

Following the XAI framework described by \cite{rizzoTheoreticalFrameworkAI2023}, an explanation is an inference made from interpreting specific evidence. In our case, we can identify $v_{t_i}$ as the evidence, with its alignment with $h_{t_i}$ providing its interpretation as corresponding to the model's conception of student proficiency. Depending on the specifics of the model's architecture, its inputs and outputs, and the parameters used, this framework would thus allow us to calculate $v_{t_i}$'s \emph{explanatory potential}.

Furthermore, following the explanation evaluation framework described by \cite{pintoUnifiedFrameworkEvaluating2024}, one of the desirable qualities of useful explanations that is enhanced when using the model's internal parameters is \emph{faithfulness}, referring to how well an explanation aligns with the model's internal state. Another important quality is \emph{plausibility}, which in our use case can be improved when students' learning trajectories align with human intuition.

\section{Discussion}

The preliminary results of our approach, as presented here, demonstrate its usefulness for interpretable knowledge tracing, particularly when using smaller datasets on which other models may overfit.\\
One promising area for future research is the exploration of alternative regularization techniques to further enhance both the performance and interpretability of the model. While our current approach uses IRT-based regularization, other models such as performance factors analysis (PFA) \cite{PavlikPFA} or BKT \cite{corbettKnowledgeTracingModeling1995} could also serve as effective regularizers. Additionally, investigating how to modify the architecture of more advanced knowledge tracing models that move away from recurrent neural networks and instead leverage mechanisms like neural attention \cite{ghosh2020contextawareattentiveknowledgetracing} or temporal point processes \cite{hawkeskt} offers an exciting path forward.

To broaden the applicability of the RegKT model, future work should focus on its scalability and computational efficiency when applied to larger and more complex datasets. While the current experiments were conducted on deliberately small datasets, real-world educational systems often involve millions of students and diverse learning tasks.

A key question is whether RegKT can maintain its interpretability while achieving even greater accuracy in such large-scale settings.

Another interesting direction is to analyze the impact of the regularizer on the model’s weight distribution in more detail, potentially uncovering deeper insights into how different regularization techniques influence model behavior.

Lastly, incorporating more complex student behaviors, such as emotional engagement and collaboration, could provide a richer understanding of the learning process \cite{baker2010better}. A potential approach is to model these behavioral data using a simple, interpretable framework and integrate it as a regularizer to capture more intricate patterns in the knowledge tracing task.

\section{Conclusion}




In this work, we introduced the RegKT model, which combines DKT and IRT to strike a balance between predictive accuracy and interpretability. While RegKT may not be the most accurate model available, it bridges the gap between two of the most widely used approaches for latent knowledge estimation: DKT and IRT.

By integrating IRT as a regularization mechanism, we mitigate the overfitting challenges typically encountered in deep learning models. At the same time, the inclusion of IRT encourages the model's weights to align more closely with interpretable signals derived from simpler statistical models.

More broadly, this approach demonstrates that using an underfitted model as a regularization tool can effectively combat overfitting, especially in data-scarce environments. This opens up interesting possibilities for improving model robustness in educational settings with limited data.

%
%

\bibliography{sigproc.bib}

\end{document}